\PassOptionsToPackage{bookmarksnumbered,unicode,hidelinks}{hyperref}

\documentclass[9pt,conference]{IEEEtran}
\IEEEoverridecommandlockouts
\usepackage{siunitx}
\usepackage{multirow}
\usepackage{bm}
\usepackage{booktabs}
\usepackage{threeparttable}
\usepackage{xcolor}
\usepackage{balance}
\usepackage{hyperref}
\usepackage{tabularx}
\usepackage{orcidlink}
\hypersetup{
    pdfauthor={},         
    pdftitle={},          
    pdfsubject={},        
    pdfkeywords={},       
    pdfproducer={},       
    pdfcreator={}         
}
\usepackage{cite}
\usepackage{amsmath,amssymb,amsfonts}
\usepackage{algorithmic}
\usepackage{graphicx}
\usepackage{textcomp}
\usepackage{xcolor}
\def\BibTeX{{\rm B\kern-.05em{\sc i\kern-.025em b}\kern-.08em
    T\kern-.1667em\lower.7ex\hbox{E}\kern-.125emX}}
\begin{document}

\title{Coflex: Enhancing HW-NAS with Sparse Gaussian Processes for Efficient and Scalable DNN Accelerator Design\\
% \thanks{\textcolor{black}{This work was supported in part by the Ministry of Education, Singapore, through its Academic Research Fund Tier 2 under Award MOE-T2EP50224-0022, and in part by SUTD through its SKI grant (SKI 2021\_02\_06). (Corresponding author: Bo Wang.)}}
\thanks{This manuscript has been accepted by the International Conference on Computer-Aided Design (ICCAD) 2025.}
}

\author{
 \IEEEauthorblockN{Yinhui Ma\IEEEauthorrefmark{1}, Tomomasa Yamasaki\IEEEauthorrefmark{1}, Zhehui Wang\IEEEauthorrefmark{2}, Tao Luo\IEEEauthorrefmark{2}, Bo Wang\IEEEauthorrefmark{1}}
\IEEEauthorblockA{\IEEEauthorrefmark{1}Singapore University of Technology and Design (SUTD), Singapore\\
\IEEEauthorrefmark{2}Institute of High Performance Computing (IHPC),\\Agency for Science, Technology and Research (A*STAR), Singapore\\
Email: \{yinhui\_ma, tomomasa\_yamasaki\}@mymail.sutd.edu.sg, \{wang\_zhehui, luo\_tao\}@ihpc.a-star.edu.sg, bo\_wang@sutd.edu.sg}
}
\maketitle

\begin{abstract}
Hardware-Aware Neural Architecture Search (HW-NAS) is an efficient approach to automatically co-optimizing neural network performance and hardware energy efficiency, making it particularly useful for the development of Deep Neural Network accelerators on the edge.
However, the extensive search space and high computational cost pose significant challenges to its practical adoption. To address these limitations, we propose Coflex, a novel HW-NAS framework that integrates the Sparse Gaussian Process (SGP) with multi-objective Bayesian optimization. By leveraging sparse inducing points, Coflex reduces the GP kernel complexity from cubic to near-linear with respect to the number of training samples, without compromising optimization performance. This enables scalable approximation of large-scale search space, substantially decreasing computational overhead while preserving high predictive accuracy. We evaluate the efficacy of Coflex across various benchmarks, focusing on accelerator-specific architecture. Our experimental results show that Coflex outperforms state-of-the-art methods in terms of network accuracy and Energy-Delay-Product, while achieving a computational speed-up ranging from 1.9$\times$ to 9.5$\times$.

%In particular, on the NATS-Bench-SSS with the ImageNet dataset, Coflex achieves a reduced error rate by 1.10\%--4.10\%, a lower Energy-Delay-Product by 98.89\%--99.85\%, and 1.5$\times$ to 11.2$\times$ faster optimization runtime compared to state-of-the-art HW-NAS methods.
\end{abstract}


\section{Introduction}
\label{sec:introduction}
\textcolor{black}{
%Recent years have witnessed a growing interest in deploying neural networks (NNs) directly on edge devices, driven by the need for real-time decision-making, reduced reliance on cloud connectivity, and increasing concerns over data privacy. 
In recent years, the increasing demand for real-time inference with enhanced data privacy has propelled the deployment of neural networks on edge devices.
Unlike cloud, edge systems impose strict constraints on computational resources and power budget. These challenges necessitate hardware-aware neural network design with co-optimization strategies that adapt to hardware constraints while maintaining excellent network performance~\cite{tanaka_pruning_2020, white_neural_2023, ozaki2022multiobjective, liu_edgeyolo_2023, quy2023edge, huang2022making, cai_once-for-all_2020}.}
%However, traditional NAS algorithms, which focus mainly on network performance, often overlook hardware constraints, resulting in inefficiencies in resource-limited applications~\cite{tanaka_pruning_2020, white_neural_2023, li_hw-nas-benchhardware-aware_2025}. 
\textcolor{black}{In this context, Hardware-Aware Neural Architecture Search (HW-NAS) has emerged as a promising approach that automatically co-optimizes neural network architectures and hardware metrics. Particularly, it can incorporate hardware constraints into the search space to jointly optimize neural network performance and hardware energy efficiency~\cite{parsa_pabo_2019, parsa2020bayesian}.} 
% \textcolor{red}{The} co-design framework further enhances efficiency by minimizing design iterations and run-time cycles.

Contemporary HW-NAS has evolved from single-objective optimization to comprehensive software–hardware multi-objective optimization.
%effectively addressing complex system-level challenges. 
%Among its recent advancements are strategies specifically tailored for In-Memory Computing (IMC), which account for process variations and hardware-level noise, significantly expanding the search space~\cite{nardi_hypermapper_2019, li_impact_2023, bhattacharjee_examining_2023}. 
%\textcolor{red}{no need to mention IMC as our paper is not based on it.}
However, the total number of \textcolor{black}{configurations} in the search space can reach up to $9.22 \times 10^{18}$ (see Table~\ref{tab:comparison-nas-bench}) even merely optimizing two objectives. %\textcolor{black}{See Table~\ref{tab:comparison-nas-bench} for an overview of parameter scale.}
%whereas optimizing more objectives can further increase the input dimension~\cite{pmlr-v180-daulton22a}. 
\textcolor{black}{This exponential growth in search complexity imposes substantial computational cost for HW-NAS.}
%particularly on energy evaluation which is typically performed using simulators such as Scale-Sim~\cite{samajdar_scale-sim_2019} and DeFiNES~\cite{mei_defines_2023}.} 
%\textcolor{red}{Need a sentence to transit between the two sentences; e.g., energy valuation is important, and usually we use simulators to get it.} 
\textcolor{black}{Moreover, incorporating hardware efficiency estimation exacerbates the computational overhead~\cite{white_neural_2023, krestinskaya2024neural}, making the framework suffer from long per-query runtime 
%(80–400 seconds) 
and frequent invocations
%often exceeding 3000 queries, required 
when exploiting a hardware simulator.} 
%\textcolor{red}{Rephrase this; the key bottleneck is not evaluation itself, but the computational costs of evaluation.}

To address these challenges, various search strategies have been proposed, such as Reinforcement Learning (RL), gradient-based optimization, and Evolutionary Algorithms (EA)~\cite{krestinskaya2024neural,blank_pymoo_2020,deb2002fast}. 
%As introduced by~\cite{krestinskaya2024neural}, RL leverages policy reuse and reward accumulation for long-term credit assignment; EA explores architectural diversity through population-level mutation and selection; and gradient-based methods relax discrete architectural choices into continuous probability distributions, often employing weight sharing to reduce training cost. 
%However, these strategies fail to demonstrate strong performance in both design space exploration and efficiency. 
However, these strategies exhibit limited effectiveness in both \textcolor{black}{search space} exploration and efficiency.
Specifically, RL is sensitive to the reward function and prone to suboptimal convergence. EA can suffer from slow optimization due to frequent simulator invocations, while gradient-based methods tend to oversimplify architecture encoding by enforcing differentiability, limiting their ability to capture discrete hardware-specific configurations. These challenges highlight a pressing need for HW-NAS frameworks that support both scalable search over vast \textcolor{black}{search space} and efficient optimization under hardware constraints. 
%To this end, we propose such a solution through the Coflex framework.} 
%Surrogate-based modeling is further enhanced by high-dimensional regression techniques such as local Gaussian Process~\cite{nguyen2009model}.} 
%\textcolor{black}{Modern HW-NAS faces increasing demand for deployment across heterogeneous task domains and datasets. A practical HW-NAS framework must generalize across tasks without requiring repeated tuning, while delivering strong trade-offs between accuracy and hardware energy efficiency~\cite{krestinskaya2024neural}.} 
%\textcolor{red}{I don't think this is what scalability means. Scalability means if the search space increases with your optimization problem, how your solution can behave.}
\textcolor{black}{To this end, we propose Coflex, a novel framework that combines Sparse Gaussian Processes (SGP)~\cite{snelson_sparse_2005} with multi-objective optimization.} 
\begin{figure*}[htbp]
\centering
\includegraphics[width=0.87\textwidth]{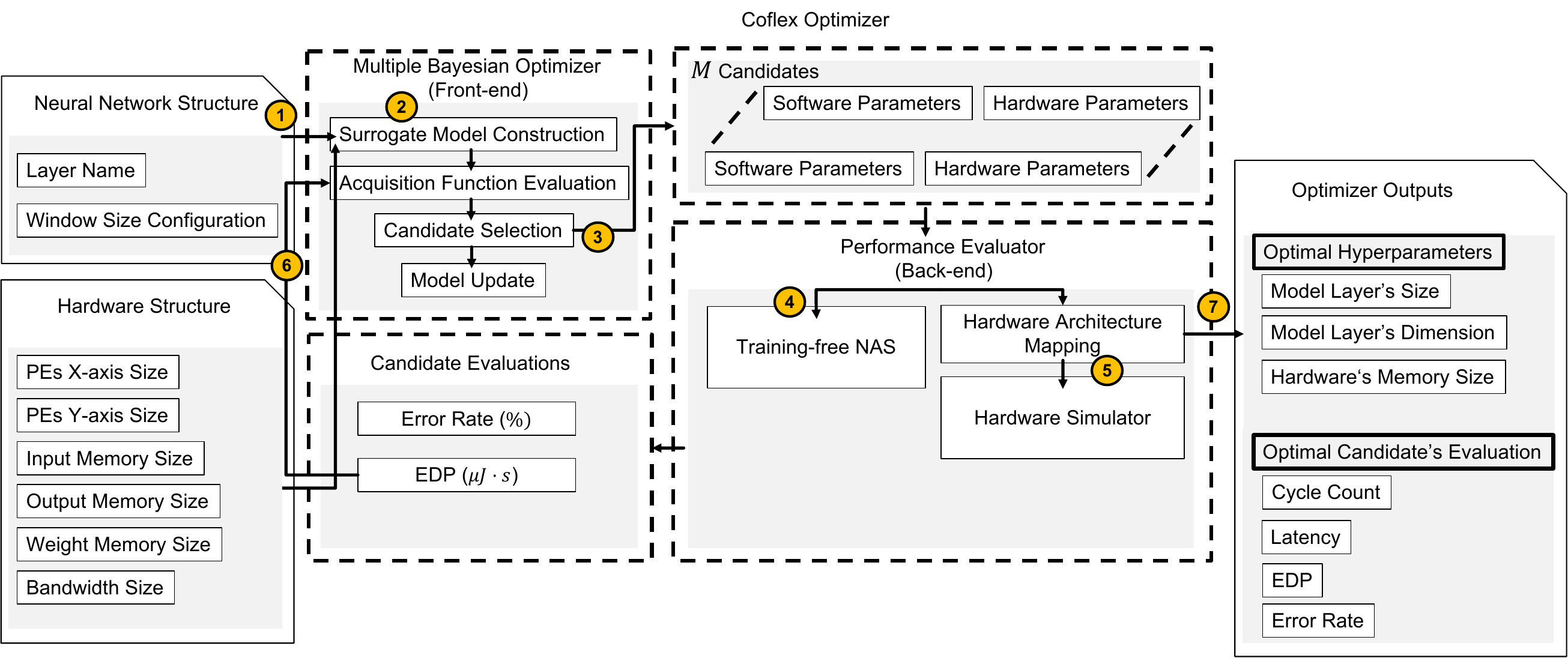}
\caption{\textcolor{black}{Overview of the Coflex framework with modular components for candidate generation, performance evaluation, and Pareto-based optimization.}}
\label{fig:hw_framework_overview}
\end{figure*}
By leveraging sparse inducing points and Bayesian optimization, Coflex approximates full Gaussian Processes (GP) with significantly lower computational costs, enabling scalable, comprehensive exploration of large search space with Pareto-optimal accuracy-energy efficiency trade-offs. 
%\textcolor{black}{In addition, Coflex has been tested across multiple types of workloads and benchmarks to validate its generalization capability in diverse applications.}
%and datasets (ImageNet, CIFAR-10/100, Taskonomy, Penn Treebank) 
%\textcolor{red}{multiple tasks imply different datasets, so we can remove the details.}
% The key contributions of our work are summarized as follows:
% \begin{itemize}
%     \item \textcolor{red}{We propose a HW-NAS framework that......}
%     \item \textcolor{black}{Coflex reduces the computational complexity of standard Gaussian Processes from $\mathcal{O}(n^3)$ to $\mathcal{O}(nm^2)$ by leveraging SGP, where $m \ll n$, while preserving model flexibility and uncertainty quantification.}
%     \item \textcolor{black}{Coflex demonstrates consistent performance across diverse NAS benchmarks and application domains, indicating strong generalization ability and scalability.}
%     \item \textcolor{black}{Coflex efficiently implements Pareto-based optimization in large-scale and high-dimensional search space, achieving robust trade-offs between neural network performance and hardware efficiency. Compared to traditional methods, Coflex exhibits superior scalability, making it well-suited for emerging deployment scenarios involving large language models (LLMs) and other future high-complexity architectures.}
% \end{itemize}

\noindent
\textcolor{black}{
The key contributions of our work are summarized as follows:
\begin{itemize}
    %\item \textcolor{black}{We introduce real-time surrogate feedback into the Bayesian optimization loop by incorporating a training-free neural performance predictor and a cycle-accurate simulator. Unlike static estimations used in prior works~\cite{parsa_pabo_2019, parsa2020bayesian}, our framework dynamically captures per-iteration objective changes, enabling posterior updates based on realistic evaluation signals and leading to higher optimization efficiency and deeper search space exploration.} \textcolor{red}{Do we need to mention Bayesian optimization? Is there any novelty involved?}
    \item We propose Coflex, a HW-NAS framework that jointly optimizes neural network performance and hardware energy efficiency for Deep Neural Network (DNN) accelerators on the edge. The framework supports Pareto-based search in large-scale search space and underscores the trade-offs between network accuracy and hardware efficiency.
    %by exploiting SGP and Bayesian optimization.
    %making it suitable for future deployment scenarios involving large language models (LLMs) and other high-complexity architectures.
    \item We integrate a sparse inducing point method to address the computational bottleneck of Gaussian Process modeling, reducing time complexity from $\mathcal{O}(n^3)$ to $\mathcal{O}(nm^2)$, where $m \ll n$ while maintaining top-1 accuracy across all evaluated benchmarks.
    \item The framework achieves optimal predictive accuracy with a computational speed-up ranging from 1.9$\times$ to 9.5$\times$ across diverse NAS benchmarks and workloads, indicating superior optimization performance and strong generalization capability.
\end{itemize}
}

\section{Architectural Overview}

\subsection{Hardware-Aware Framework}
\label{sec:intro_hw_nas_framework}

Coflex is an HW-NAS framework based on SGP that efficiently explores large-scale, multidimensional search space. \textcolor{black}{As Figure~\ref{fig:hw_framework_overview} shows, the input consists of both neural network configurations (e.g., layer type, window size) and hardware configurations (e.g., memory size, PE array layout) (Step~1). In each optimization iteration, the front-end module constructs sparse surrogate models for error rate and Energy-Dealy-Product (EDP) using SGP (Step~2). Acquisition Function (AF) evaluation is performed to select promising candidates from the search space, which are subsequently forwarded to the back-end evaluator (Step~3). The back-end integrates a training-free NAS performance evaluator (Step~4) and a cycle-accurate DNN accelerator simulator (Step~5) to compute realistic objective values for each candidate. The back-end then returns these feedback signals to update the surrogate model in the front-end (Step~6). Through iterative optimization, Coflex refines its search and progressively converges towards Pareto-optimal design configurations. In Step~7, the final output is generated, including optimal neural network architecture and hardware configurations, and the performance metrics, such as error rate, latency, EDP, and cycle count.}

% (use active voice, and better to enrich below to elaborate Fig. 1 in detail)} Coflex uniformly samples the search space using Latin Hypercube Sampling (LHS), subsequently decomposes it by dimension, followed by optimizing each subspace with single-objective GP. The outputs are concatenated and pruned via Pareto non-dominance filtering, yielding a sparse covariance approximation that enables the multi-objective GP to converge to high-quality candidate solutions rapidly. 

%\textcolor{red}{which will be elaborated in Section~\ref{sec:experimental_results}}.
%(See in Section V)
 \textcolor{black}{Figure~\ref{fig:dimension_decomposation} illustrates the end-to-end workflow of Coflex. The pipeline begins with initial sample generation (Step~1), followed by dimension decomposition (Step~2), which separates the joint configuration space into two orthogonal subspaces: energy-wise hyperparameters and error-wise hyperparameters. For each subspace, initial candidates are evaluated separately. A cycle-accurate DNN accelerator simulator~\cite{mei_defines_2023} is exploited to estimate energy efficiency (Step~3), while a training-free NAS algorithm~\cite{rbflexnas2025} is utilized to predict error rate. These evaluations generate scalar feedback on EDP and error rate to network models, respectively. These models serve as posterior distributions for AF optimization (Step~5), where Coflex selects the next query points that maximize a weighted combination of acquisition values from each objective. \textcolor{black}{In Step 6, the selected candidate $\mathbf{x}_{\text{next}}$ is re-evaluated to obtain updated objective values. These observations are then appended to the datasets for each objective, enabling posterior refinement in subsequent steps.} To ensure the quality of multi-objective trade-off analysis, the Pareto-Front Filtering module (Step~7) eliminates dominated candidates and enforces diversity, followed by a refinement step (Step~8) to filter corresponding observations for posterior updates. To enhance scalability for large search space, Coflex incorporates inter-part posterior fusion (Step~9), enabling knowledge transfer between energy-wise and error-wise GPs without incurring full kernel cost. Step~10 extracts the best-performing candidates on the current Pareto front, while Step~11 reinitializes the sampling pool based on updated priors. These steps complete a full Coflex optimization round before entering the next iteration.}

% illustrates XXX (enrich below to elaborate Fig. 2 in detail)}
% Specifically, each batch of optimization evaluates $M$ configurations incorporating software and hardware configurations through a state-of-the-art (SOTA) training-free NAS algorithm ~\cite{rbflexnas2025}
% %RBFlex-NAS~\cite{rbflexnas2025}
% and a cycle-accurate hardware simulator~\cite{mei_defines_2023}.
% %DeFiNES~\cite{mei_defines_2023}. 
% The resulting objectives are used to update the Bayesian optimizer and guide acquisition refinement. Upon convergence, Coflex yields a configuration that approximates a near-optimal solution for both neural network performance and hardware energy efficiency, which eventually facilitates automated software-hardware co-design.

\subsection{Search Space Formulation}
\label{sec:design_space_define}

Our HW-NAS search space incorporates both neural network and hardware constraints to enable optimization for neural network performance and hardware energy efficiency simultaneously.

\subsubsection{Joint Search Space}
\label{sec:software_framework_design_space}

\begin{figure}[htbp]
  \centering
  \includegraphics[width=0.7\linewidth]{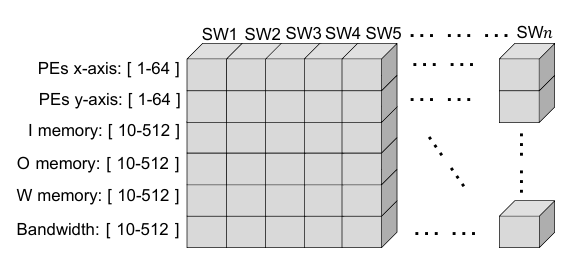}
  \caption{\textcolor{black}{Illustration of a NAS benchmark-driven, hardware-aware search space. The number of software configuration $n$ is defined as $n=5$ for NATS-Bench-SSS, $n=6$ for TransNAS-Bench-101, and $n=8$ for NAS-Bench-NLP.}}
  \label{fig:design_space_of_nas_ss}
\end{figure}

\textcolor{black}{Network topology and model size are widely recognized as primary factors that impact neural network performance~\cite{white_neural_2023}.} 
\textcolor{black}{Common NAS benchmarks, such as NATS-Bench-SSS \cite{dong_nats-bench_2021} and NAS-Bench-201 \cite{dong_nas-bench-201_2020} 
leverage these \textcolor{black}{configurations to systematically explore search space.}}
%This relationship has been explicitly validated by benchmarks including NATS-Bench \cite{dong_nats-bench_2021} and NAS-Bench-201 \cite{dong_nas-bench-201_2020}, and recent NAS surveys \cite{white_neural_2023}, which quantify the effect of structural variation and parameter scaling on accuracy.} \textcolor{red}{Any paper to prove it?} 
For instance, the NATS-Bench-SSS framework~\cite{dong_nats-bench_2021} provides a standardized search space comprising 15,625 candidate topologies and 32,768 distinct sizes. In addition to topology and size, the software search space includes layer types, activation functions, normalization methods, and connection strategies. It also incorporates configurations such as the number of filters, kernel sizes, and learning rates. 
%Specifically for Workload Type 1 \textcolor{red}{in Section~\ref{sec:experimental_results}}, Coflex encodes each neural network candidate as a five-dimensional vector, with each dimension corresponding to a convolutional layer size of $8 \times n$, where $n \in \{1, 2, \dots, 8\}$.
\begin{figure*}[htbp]
\centering
\includegraphics[width=0.78\textwidth]{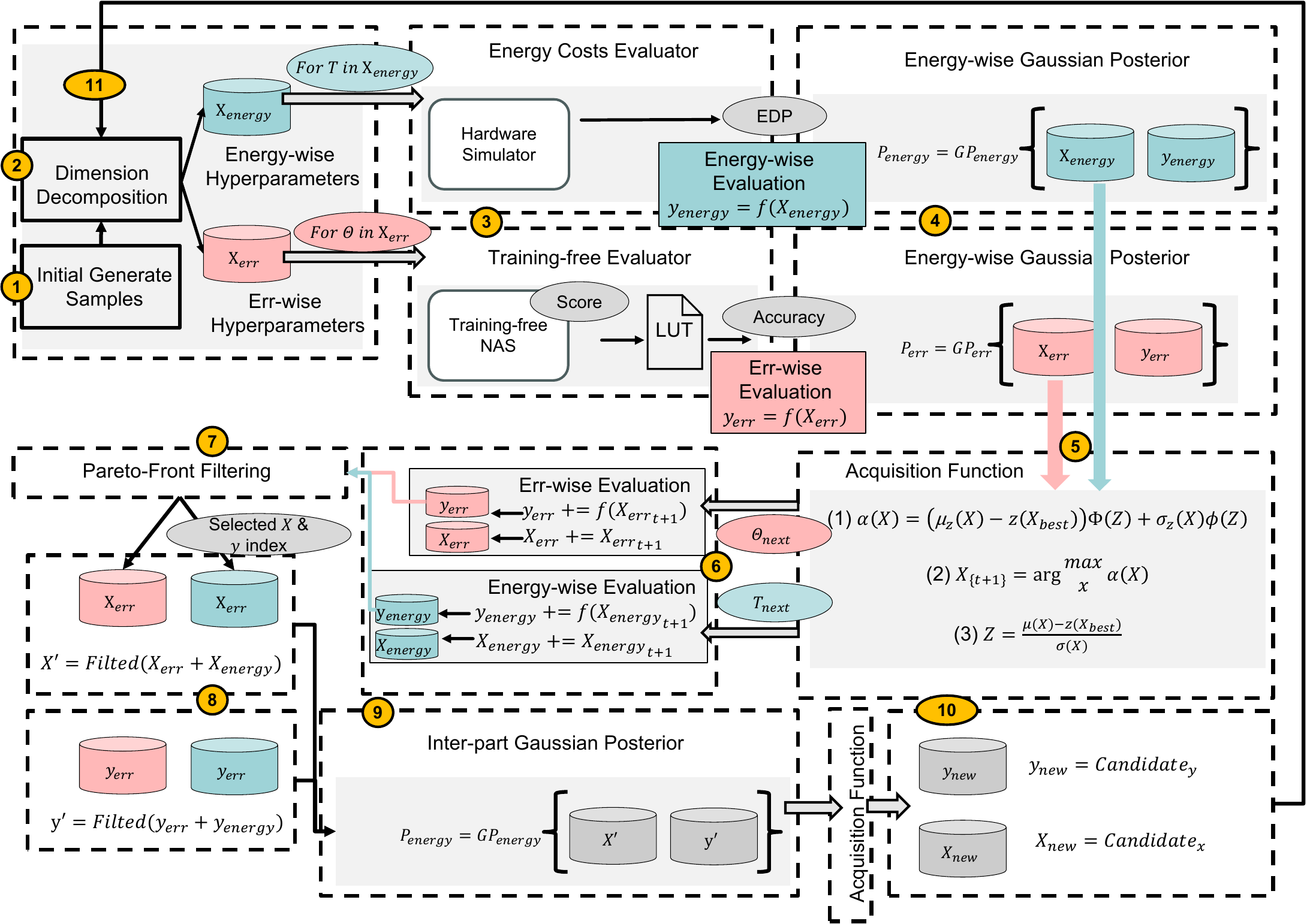}
\caption{\textcolor{black}{Illustration of Coflex with dimension decomposition and sparse Gaussian inducing strategies.}}
\label{fig:dimension_decomposation}
\end{figure*}
\textcolor{black}{On the hardware side, Coflex spans six dimensions, including the number of processing elements (PEs) along x- and y-axes, input memory capacity (\(I_\text{mem}\)), output memory capacity (\(O_\text{mem}\)), weight memory capacity (\(W_\text{mem}\)), and on-chip bus bandwidth (Figure~\ref{fig:design_space_of_nas_ss}).} 
Specifically, the PE count is adjustable from 1 to 10, while the memory sizes and bandwidth can be set to any integer between 64 and 512. 
The joint search space integrates both the software and hardware search spaces for co-optimization.
%This joint software-hardware search space enables the co-optimization of neural network performance and hardware energy efficiency.
%providing the flexibility required to adapt HW-NAS to heterogeneous platforms.

\subsubsection{Initial Data Sampling}
\label{sec:initial data sampling}

Due to the complexity of HW-NAS search space, it is essential to perform \textcolor{black}{uniform sampling for initial data from search space}. Sobol sampling and Latin Hypercube Sampling (LHS) are common uniform sampling approaches in high dimensions. \textcolor{black}{The former produces a quasi-random, low-discrepancy sequence to cover the sample space more uniformly than random sampling. \textcolor{black}{However, it can exhibit clustering or uneven coverage when the \textcolor{black}{sample data points} are limited (i.e., fewer than 100).}
In contrast, the latter is a stratified sampling method that ensures each dimension is evenly divided into intervals, achieving better coverage under limited sample data.
Therefore, we adopt LHS in the Coflex framework for initial data sampling in high-dimensional search space.} 
%\textcolor{red}{briefly introduce what Sobol sampling is and what LHS is.}

\section{Sparse Gaussian-Based Optimization}
\label{sec:sparse_gaussian_based_optimization}

\subsection{Gaussian Process Basics}
\label{sec:gaussian_process_basic}

Gaussian process~\cite{mackay1998introduction} provides a non-parametric Bayesian framework for modeling complex objective functions in HW-NAS, allowing closed-form posterior inference under Gaussian assumptions. Specifically, for a given training dataset
\begin{itemize}
    \item $X = \{x_1, \dots, x_n\} \in \mathbb{R}^{n \times d}$ \textcolor{black}{where $X$ is} the set of $n$ input points in $d$ dimensions;
    \item $y = [y_1, \dots, y_n]^\top \in \mathbb{R}^n$ \textcolor{black}{where $y$} corresponds to observed values, where each $y_i = f(x_i) + \epsilon_i$, and $\epsilon_i \sim \mathcal{N}(0, \sigma^2)$ represents Gaussian observation noise. 
\end{itemize}
\noindent
As we wish to predict the function output $y_\ast$ at a new test point $x_\ast$, we then assume the objective function $f(x)$ satisfies a GP prior as
\begin{equation}
f(x) \sim \mathcal{GP}(\mu(x), k(x, x'))
\label{eq:gp_prior}
\end{equation}
\noindent
The function values at any finite set of inputs follow a multivariate Gaussian distribution as

\begin{equation}
\begin{bmatrix}
y \\
f_*
\end{bmatrix}
\sim \mathcal{N} \left(\mu(x), 
\begin{bmatrix}
K(X,X) + \sigma^2 I & K(X, x_*) \\
K(x_*, X) & K(x_*, x_*)
\end{bmatrix}
\right)
\label{eq:gp_joint}
\end{equation}

\noindent
where
\begin{itemize}
    \item $K(X, X) \in \mathbb{R}^{n \times n}$ \textcolor{black}{where $K(X, X)$ is a} covariance matrix between training points (including noise);
    \item $K(X, x_*) \in \mathbb{R}^{n \times 1}$ \textcolor{black}{where $K(X, x_*)$ represents} covariance between training points and test point;
    \item $K(x_*, x_*) \in \mathbb{R}$ \textcolor{black}{where $K(x_*, x_*)$ is the} variance at the test point.
\end{itemize}

\noindent
According to the conditional distribution of the multivariate Gaussian, we obtain the equations as follows.
\begin{equation}
\mu_*(x_*) = K(x_*, X)\left[K(X, X) + \sigma^2 I\right]^{-1} y
\label{eq:gp_posterior_mean}
\end{equation}
\begin{equation}
\sigma^2_*(x_*) = K(x_*, x_*) - K(x_*, X)\left[K(X, X) + \sigma^2 I\right]^{-1} K(X, x_*)
\label{eq:gp_posterior_var}
\end{equation}
\noindent
\textcolor{black}{Here, $\mu_*(x_*)$ denotes the predicted mean output at the test point $x_*$, and 
$\sigma^2_*(x_*)$ is the predictive uncertainty.}
%These correspond to the first and second moments of the Gaussian process posterior, i.e.,}
\iffalse
\begin{equation}
\mu_*(x_*) = \mathbb{E}[f(x_*) \mid X, y]
\label{eq:gp_mean_closed_form}
\end{equation}
\begin{equation}
\sigma^2_*(x_*) = \mathrm{Var}[f(x_*) \mid X, y]
\label{eq:gp_var_closed_form}
\end{equation}
\fi

\noindent
%\textcolor{red}{what sigma is?} 
In this work, we adopt Gaussian Process for multi-objective optimization. Particularly, we scalarize the objectives using weight vectors, and guide the search by 
%The objectives are scalarized using weight vectors, and the search is guided by 
     Acquisition Functions (AF), including Expected Improvement (EI)~\cite{qin_improving_2017}, which satisfies
\begin{equation}
\text{EI}(x) = \mathbb{E} \left[ \max(0, f(x) - f_{\text{best}}) \right]
\label{eq:ei_function}
\end{equation}

\noindent
where
\[
f(x_*) \sim \mathcal{N}(\mu_*(x_*), \sigma^2_*(x_*))
\]

\noindent
The next query point is chosen by maximizing the acquisition value as
\begin{equation}
x_{\text{next}} = \arg\max_{x \in \mathcal{X}} \mathrm{EI}(x)
\end{equation}

\textcolor{black}{Our Bayesian optimization framework, which leverages GP solely as surrogate models, preserves non-dominated solutions in each iteration by maintaining Pareto diversity and mitigating GP model bias.} 
In order to alleviate the cubic complexity which is inherent to standard GP, SGP with inducing points is employed, reducing time complexity from \( O(n^3) \) to \( O(m^2 n) \) where \( m \ll n \). This approximation supports scalable inference over large-scale, high-dimensional search space, enabling effective analysis of trade-offs between multiple objectives.

\begin{table*}[htbp]
\centering
\caption{Various Workloads for Coflex Evaluation}
%\textcolor{red}{Do the bench suits include HW parameters to explore?}
\label{tab:comparison-nas-bench}
\resizebox{0.75\textwidth}{!}{%
\begin{tabular}{lccc}
    \toprule
    \textbf{} & \textbf{Type 1 Workload} & \textbf{Type 2 Workload} & \textbf{Type 3 Workload} \\
    \midrule
    Neural Network Benchmark     & NATS-Bench-SSS~\cite{dong_nats-bench_2021}   & TransNAS-Bench-101~\cite{duan_transnas-bench-101_2021}   & NAS-Bench-NLP~\cite{klyuchnikov2022bench} \\
    SW Search Space     & $3.2 \times 10^{4}$     & $4.1 \times 10^{3}$     & $1.43 \times 10^{4}$ \\
    HW Search Space (w.r.t. simulated architecture)     & $2.81 \times 10^{14}$   & $2.81 \times 10^{14}$   & $2.02 \times 10^{15}$ \\
    Total \textcolor{black}{Configurations}    & $9.22 \times 10^{18}$   & $1.15 \times 10^{18}$   & $2.89 \times 10^{19}$ \\
    \bottomrule
\end{tabular}%
-}
% \begin{tablenotes}
% \item \textit{\textcolor{black}{Note that the HW search space is task-adaptive and varies with each benchmark to reflect different deployment needs.} \textcolor{red}{Not sure what it means.}}
% \end{tablenotes}
\end{table*}

\subsection{Structured Decomposition for Multi-Objective Optimization}
\label{sec:dimension_decomposition}

Multi-objective Bayesian optimization in high-dimensional search space often suffers from prohibitive computational complexity and degraded model fidelity, particularly when multiple trade-offs must be maintained simultaneously. To address this challenge, prior work~\cite{parsa_pabo_2019, parsa2020bayesian} introduced a hierarchical pseudo-agent-based approach that enables scalable surrogate modeling. \textcolor{black}{Coflex builds upon this prior hierarchical decomposition strategy but integrates SGP into a novel two-level surrogate architecture, which explicitly decouples software and hardware objectives. As Table~\ref{tab:comparison-image} shows, this integration enables more efficient modeling, which drives the Pareto front closer to the reference point under constrained evaluation budgets than prior work.} 
%\textcolor{red}{This sentence seems we are imitating prior art. I think we need also highlight our difference or innovation} 

Specifically, the multi-objective problem is first decomposed into independent single-objective subproblems. This allows each Gaussian Process to operate over a reduced-dimensional subspace, significantly reducing the size of its covariance matrix. A supervisor Gaussian process then aggregates the outputs and constructs a unified surrogate model to steer the Pareto front toward a specified reference point. While dimension decomposition reduces individual Gaussian model complexity, the final aggregation still produces a large covariance matrix, which complicates the synthesis of globally optimal candidate solutions. Section~\ref{sec:algorithmic_time_complexity_measurement} further discusses this issue and provides strategies to manage high-dimensional inputs.

\subsection{Sparse Gaussian Inducing Strategies}
\label{sec:sparse_gaussian_inducing_strategies}

%\textcolor{red}{This section is very long and lacks sufficient figures; Simplifiy it with figures for better illustration}
\textcolor{black}{
As Figure~\ref{fig:dimension_decomposation} illustrates, the inter-part posterior fusion process in Step~9 still necessitates the inversion of large-scale covariance matrices (Eqs.~\eqref{eq:gp_posterior_mean} and~\eqref{eq:gp_posterior_var}) when solely relying on dimension decomposition. This introduces significant computational overhead, limiting the scalability of the approach.}
\textcolor{black}{
To further reduce computational complexity while preserving high-quality optimization, Coflex adopts a sparse inducing point strategy. This is because the selection of inducing points plays a critical role in determining the fidelity of posterior approximation in SGP. To ensure that the most informative samples are retained, Coflex employs Pareto front filtering to extract high-quality, non-dominated candidates from the concatenated search space. Specifically, a solution \( \mathbf{x}_i \) is deemed non-dominated if no other solution \( \mathbf{x}_j \in \mathcal{X} \) satisfies:
\begin{equation}
\begin{split}
f_k(\mathbf{x}_j) \leq f_k(\mathbf{x}_i), \quad \forall k \in \{1, \dots, K\}, \\
\text{and } f_k(\mathbf{x}_j) < f_k(\mathbf{x}_i) \, \text{for at least one } k,
\end{split}
\label{eq:dominate_checking}
\end{equation}
where \( f_k \) denotes the \( k \)-th objective function, and \( K \) is the total number of objectives. These candidates, residing near the boundaries of multi-objective optima, provide structurally representative anchors for posterior construction.
} 

\begin{table}[htbp]
\caption{Variable definitions for the RNN energy and latency estimation}
\label{tab:rnn_energy_latency_vars}
\centering
\begin{tabular}{l p{6.8cm}}
\toprule
$M$ & Number of operation types \\
$E_{\text{Total}}$ & Total estimated energy consumption (Joules) \\
$L_{\text{Total}}$ & Total estimated execution latency (seconds) \\
$E_{\text{op},i}$ & Energy per operation at frequency $f_i$ \\
$E_{\text{ref},i}$ & Reference energy per operation at $f_{\text{ref}}$ \\
$f_i$ & Clock frequency for operation type $i$ (MHz) \\
$f_{\text{ref}}$ & Reference clock frequency (MHz) \\
$\beta$ & Frequency-energy exponent, typically in $[0.5, 2.0]$ \\
$d_i$ & Input data dimension (e.g., hidden size) \\
$n_i$ & Number of operations per data unit or number of PEs used \\
$n_{\mathrm{PE},i}$ & Number of parallel processing elements for operation $i$ \\
$\alpha_i$ & Process node scaling factor, e.g., $\alpha_i = 28/\text{node (nm)}$ \\
$\eta_i$ & Efficiency multiplier for PE utilization and memory (range: 0.0–1.0) \\
\bottomrule
\end{tabular}
\end{table}

To realize the sparse covariance structure, Coflex adopts the Matérn kernel~\cite{seeger_gaussian_2004} to quantify pairwise similarities between data points and inducing anchors. The Matérn kernel is defined as:

\begin{equation}
k_{\nu}(r) = \sigma^2 \cdot \frac{2^{1-\nu}}{\Gamma(\nu)} 
\left( \sqrt{2\nu} \cdot \frac{r}{\ell} \right)^{\nu} 
K_{\nu}\left( \sqrt{2\nu} \cdot \frac{r}{\ell} \right)
\label{eq:matern_general}
\end{equation}

\noindent
where $r = \| \mathbf{x} - \mathbf{x}' \|$ is the Euclidean distance, 
$\nu$ controls the smoothness, $\ell$ is the length scale, and 
$K_\nu(\cdot)$ denotes the modified Bessel function of the second kind.

We use the common setting $\nu = 3/2$, yielding a simplified form:

\begin{equation}
k_{3/2}(r) = \sigma^2 \left(1 + \frac{\sqrt{3}r}{\ell} \right) 
\exp\left(-\frac{\sqrt{3}r}{\ell} \right)
\label{eq:matern_32}
\end{equation}
By applying this kernel, Coflex computes the following covariance matrices:
\begin{itemize}
  \item Cross-covariance: $K_{XZ} \in \mathbb{R}^{n \times m}$ between training and inducing points
  \item Inducing-point covariance: $K_{ZZ} \in \mathbb{R}^{m \times m}$
  \item Transposed counterpart: $K_{ZX} = K_{XZ}^\top$
\end{itemize}

\textcolor{black}{
These matrices form a low-rank decomposition tuple $(U = K_{XZ}, C = K_{ZZ}^{-1}, V = K_{ZX})$, which directly supports fast posterior inference via the Woodbury identity in Section~\ref{sec:algorithmic_time_complexity_measurement}. It bridges the geometry of structural space into kernel space, enabling Coflex to reduce computational overload while maintaining posterior fidelity.
}

\subsection{Time Complexity Reduction}
\label{sec:algorithmic_time_complexity_measurement}

\textcolor{black}{Standard Gaussian Process regression constructs an \( n \times n \) covariance matrix \( K_{XX} \) from \( n \) training points, 
as derived in Eq.~(\ref{eq:gram_matrix}). Inverting this matrix typically incurs a computational cost of \( \mathcal{O}(n^3) \) via Cholesky factorization~\cite{cholesky_sur_2005}. Here, \( K_{XX} \in \mathbb{R}^{n \times n} \) denotes the Gram matrix, computed over the training inputs \( X = \{x_1, x_2, \ldots, x_n\} \subset \mathbb{R}^d \). Each entry \( k(x_i, x_j) \) represents the kernel function evaluating the similarity between the input pairs \( (x_i, x_j) \).}
\begin{equation}
K_{XX} =
\begin{bmatrix}
k(x_1, x_1) & k(x_1, x_2) & \cdots & k(x_1, x_n) \\
k(x_2, x_1) & k(x_2, x_2) & \cdots & k(x_2, x_n) \\
\vdots      & \vdots      & \ddots & \vdots      \\
k(x_n, x_1) & k(x_n, x_2) & \cdots & k(x_n, x_n)
\end{bmatrix}
\label{eq:gram_matrix}
\end{equation}

\textcolor{black}{Similarly, we obtain the covariance matrices $K_{XZ}$, $K_{ZX}$, and $K_{ZZ}$ using the Matérn kernel described in Section~\ref{sec:sparse_gaussian_inducing_strategies} to quantify pairwise similarities between training points and inducing points. }
\textcolor{black}{In Sparse GP, instead of directly modeling the full covariance matrix \( K_{XX} \in \mathbb{R}^{n \times n} \), a smaller set of inducing points \( Z \in \mathbb{R}^{m \times d} \), where \( m \ll n \), is introduced to approximate the full kernel.}
%as shown in Eq.~(\ref{eq:fitc_approx}).} 
%The following covariance blocks are subsequently constructed \textcolor{red}{$K_{XZ}$, $K_{ZX}$, and $K_{ZZ}$.} 
% The former is an $n \times m$ cross-covariance between the original data $X$ and the inducing points $Z$ and the latter is an $m \times m$ covariance matrix among the inducing points alone. 

Specifically, the matrix inversion lemma, also known as the Woodbury identity, provides a means to efficiently compute the inverse of a low-rank update (i.e., $\mathbf{U}\,\mathbf{C}\,\mathbf{V}$) to the base matrix $\mathbf{A}$ as follows:
\begin{equation}
(\mathbf{A} + \mathbf{U}\,\mathbf{C}\,\mathbf{V})^{-1} 
\;=\; 
\mathbf{A}^{-1}
\;-\; 
\mathbf{A}^{-1}\,\mathbf{U}\,\Bigl(\mathbf{C}^{-1} + \mathbf{V}\,\mathbf{A}^{-1}\,\mathbf{U}\Bigr)^{-1}\,\mathbf{V}\,\mathbf{A}^{-1}
\label{eq:woodbury_identity}
\end{equation}
\begin{equation}
\mathbf{A} = \sigma^2 \mathbf{I} \in \mathbb{R}^{n \times n}
\label{eq:noise_section}
\end{equation}
where $\mathbf{A}$ is chosen as a large, easily invertible matrix (e.g. $n \times n$).
%while $\mathbf{U}\,\mathbf{C}\,\mathbf{V}$ represents a \textcolor{black}{low-rank update to the base matrix $\mathbf{A}$.} 
By expressing a large matrix \( \mathbf{A} + \mathbf{U}\,\mathbf{C}\,\mathbf{V} \) in this form, the Woodbury identity is used to avoid the full \( \mathcal{O}(n^3) \) matrix inversion in Eqs.~\eqref{eq:gp_posterior_mean} and~\eqref{eq:gp_posterior_var}, but focus on the much smaller \( \mathbf{C}^{-1} \) and additional low-rank multiplications instead, substantially reducing computational cost. 

However, the observed training outputs $y$ do not directly reflect the latent function $f(x)$, but are corrupted by noise as shown in Section~\ref{sec:gaussian_process_basic}. To incorporate this noise into the GP model, the covariance matrix is adjusted as
\begin{equation}
K = K_{XX} + \sigma^2 I
\end{equation}
When the full GP covariance $K_{XX}$ is represented in terms of $K_{XZ}$, $K_{ZZ}$, and $K_{ZX}$, the approximation applies as follows:
\begin{equation}
K_{XX} \approx K_{XZ}\,K_{ZZ}^{-1}\,K_{ZX} + \operatorname{diag}\Bigl(K_{XX} - K_{XZ}\,K_{ZZ}^{-1}\,K_{ZX}\Bigr)
\label{eq:fitc_approx}
\end{equation}
\begin{equation}
K_{XX} + \sigma^2 I \approx K_{XZ} K_{ZZ}^{-1} K_{ZX} + \sigma^2 I
\label{eq:approx}
\end{equation}
where
\begin{itemize}
    \item \( K_{XZ} \in \mathbb{R}^{n \times m} \) is the cross-covariance between training points \( X \) and inducing points \( Z \), corresponding to $\mathbf{U}$ in the Woodbury dentity;
    \item \( K_{ZZ} \in \mathbb{R}^{m \times m} \) is the covariance between inducing points, corresponding to $\mathbf{C}$;
    \item \( K_{ZX} = K_{XZ}^{\top} \), corresponding to $\mathbf{V}$.
\end{itemize}
\textcolor{black}{We apply the Woodbury identity by letting $\mathbf{A} = \sigma^2 \mathbf{I}$, $\mathbf{U} = K_{XZ}$, $\mathbf{C} = K_{ZZ}^{-1}$, and $\mathbf{V} = K_{ZX}$. Under this formulation, the inversion of the covariance matrix $(K_{XX} + \sigma^2 I)^{-1}$ can be computed via the Woodbury identity, eliminating the need for directly inverting an $n \times n$ matrix, which otherwise incurs a computational cost of $\mathcal{O}(n^3)$. Computing $K_{ZZ}^{-1}$ requires $\mathcal{O}(m^3)$ time, where $m$ is the number of inducing points, and the matrix product $K_{XZ} K_{ZZ}^{-1} K_{ZX}$ costs $\mathcal{O}(nm^2)$. Therefore, the overall time complexity becomes $\mathcal{O}(nm^2 + m^3)$. When $m \ll n$, the dominant term is $\mathcal{O}(nm^2)$, providing near-linear scalability with respect to the number of training samples.}

\section{Evaluation Approach}

\subsection{Training-Free Neural Architecture Evaluation}
\label{sec:rbflex_nas}

We adopt RBFlex-NAS~\cite{rbflexnas2025}, a state-of-the-art NAS algorithm for network performance evaluation in our framework.
%RBFlex-NAS~\cite{rbflexnas2025} serves as the software-side performance evaluator within the framework. 
The method assesses candidate solutions from the multi-objective Gaussian process in a training-free manner~\cite{mellor_neural_2021}. 
Specifically, the evaluation measures similarity between activation outputs and similarity between final-layer input features across multiple input samples. Through this approach, it delivers superior accuracy with minimal computational cost, outperforming other training-free methods, including TAS~\cite{dong_network_2019}, FBNet-v2~\cite{NAS_FBNet}, TE-NAS~\cite{chen_neural_2021}, NASWOT~\cite{mellor_neural_2021}, and ZiCo~\cite{li_zico_2023}. Additionally, it supports a variety of activation functions, allowing its application to a wider range of tasks, including image classification, semantic segmentation, and natural language processing.
%In NLP benchmarks, for example, NASWOT often underperforms due to its limited support for diverse activation layers, despite offering comparable search efficiency.

\subsection{Evaluation for Hardware Prototype}
% \label{sec:defines}
\label{sec:evaluation_for_hardware_deployment}

\begin{figure}[!t]
  \centering
  \includegraphics[width=\linewidth]{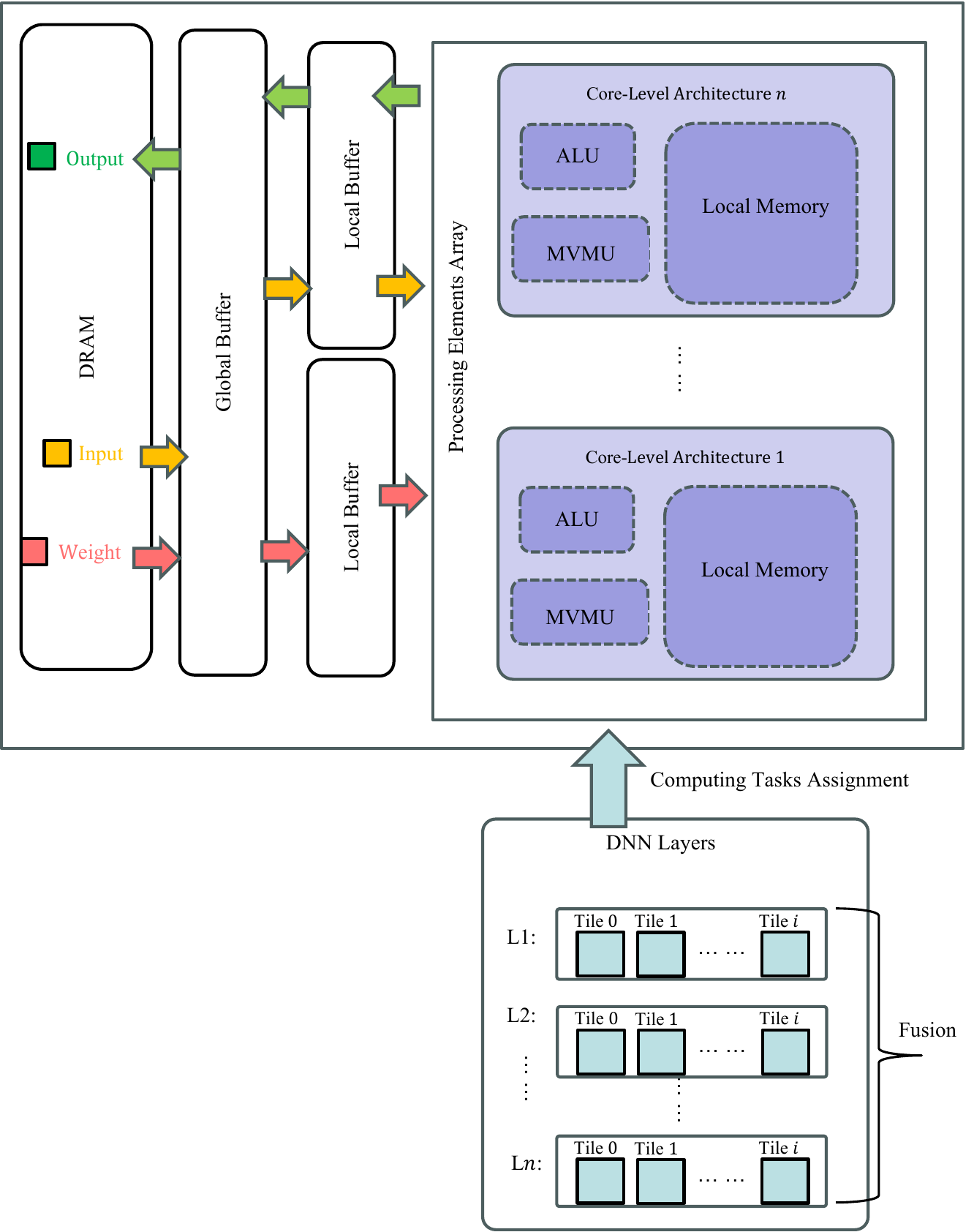}
  \caption{Overview of the accelerator prototype for hardware space search.}
  \label{fig:def_arch}
\end{figure}

We integrate DeFiNES~\cite{mei_defines_2023}, a cycle-accurate DNN accelerator simulator to estimate the hardware metrics of candidate solutions when they are deployed on a tile-based, Meta-proto-like architecture ~\cite{meta-proto}.
DeFiNES models this architecture (Figure~\ref{fig:def_arch}) by spatially partitioning DNN layers into tiles, which are then dispatched to PEs. Each PE comprises a Matrix-Vector Multiply Unit (MVMU), an Arithmetic Logic Unit (ALU), and local memory for efficient computation. A hierarchical memory system coordinates the movement of inputs, weights, and outputs to minimize energy and latency.
%\textcolor{red}{show details of underlying hardware architecture with a reference}
%DeFiNES~\cite{mei_defines_2023} is employed as a real-time evaluator of hardware energy efficiency for candidate solutions generated by the multi-objective Gaussian Process.

The predictions of the simulator for latency and energy consumption have been quantitatively validated against actual hardware-level measurements~\cite{mei_defines_2023}, using a variety of neural network models. Experimental results show that it achieves an average latency prediction error of 3\%, with a worst-case error of 10\% for FSRCNN while energy estimation error remains within 6\%. The results validate that DeFiNES can effectively capture hardware metrics, enabling rapid evaluation without the need for physical hardware implementation.

%\subsubsection{Evaluation for RNN Deployment}
%\label{sec:rnn}
\textcolor{black}{The NAS benchmarks employed in this work primarily target neural network architectures based on Convolutional Neural Networks (CNN) and Recurrent Neural Networks (RNN). However, the DeFiNES simulator is limited to supporting CNN-based architectures only.
As it can not evaluate the energy or latency of RNN candidates from NAS benchmarks,} 
%Therefore, an alternative evaluation methodology is required.
we adopt a method that has been widely utilized~\cite{azari2020elsa,li_e-rnn_2019,silfa_e-pur_2018,paulin2021vau} to estimate RNN models. 
%E-RNN~\cite{li_e-rnn_2019}, E-PUR~\cite{silfa_e-pur_2018}, and Vau da Muntanialas~\cite{paulin2021vau}.
Specifically, the total energy consumption $E_{\text{Total}}$ and latency $L_{\text{Total}}$ are estimated by aggregating the contributions of all operations involved
\begin{table}[htbp]
\centering
\caption{\textcolor{black}{NAS Evaluation Settings for Type 1 Workload (NATS-Bench-SSS)}}
\label{tab:exp-nas}
\resizebox{0.45\textwidth}{!}{%
\begin{tabular}{ll}
\toprule
\textbf{Items} & \textbf{NAS} \\
\midrule
\textbf{Dataset} & ImageNet / CIFAR100 / CIFAR10 \\
\textbf{Network Benchmark} & NATS-Bench-SSS \\
\textbf{Input Image Size} & (224,224), (32,32) \\
\textbf{Iters} & 30 \\
\textbf{Initial Samples} & 100 \\
\textbf{All Algorithms Tested} & Yes \\
\textbf{Mapping} & Output layer \\
\textbf{Layer Types} & Activation Layers \\
\textbf{HW Optimization Configs} & 
\begin{tabular}[t]{@{}l@{}}
  PE-x, PE-y: 1--64 \\
  Mem-I: 10--512 \\
  Mem-O: 10--512 \\
  Mem-W: 10--512 \\
  Bandwidth: 10--512
\end{tabular} \\
\bottomrule
\end{tabular}%
}
\end{table}
\begin{table}[htbp]
\centering
\caption{\textcolor{black}{Simulator Configuration: Meta-Prototype Architecture for HW Evaluation}}
\label{tab:exp-sim}
\resizebox{0.45\textwidth}{!}{%
\begin{tabular}{ll}
\toprule
\textbf{Items} & \textbf{Simulator} \\
\midrule
\textbf{Hardware Architecture} & Meta-prototype-DF-like \\
\textbf{Registers / LBs / GB} & Reconfigurable \\
\textbf{Data Storing Modes} & Fully-cached \\
\textbf{Supported Layers} & Conv / BN / ReLU / Maxpool \\
\textbf{Tilesize-x} & 32 \\
\textbf{Tilesize-y} & 32 \\
\bottomrule
\end{tabular}%
}
\end{table}
\noindent
in the RNN computation graph. 
Specifically, for each operation type $i$, the energy and latency calculations are given by
\begin{align}
E_{\text{Total}} &= \sum_{i=1}^{M} E_{\text{op},i} \cdot d_i \cdot n_i \cdot \alpha_i \cdot \eta_i \notag \\
&= \sum_{i=1}^{M} E_{\text{ref},i} \cdot \left( \frac{f_{\text{ref}}}{f_i} \right)^{\beta} \cdot d_i \cdot n_i \cdot \alpha_i \cdot \eta_i
\label{eq:energy_final}
\end{align}
\begin{equation}
L_{\text{Total}} = \sum_{i=1}^{M} \frac{d_i \cdot n_i}{n_{\mathrm{PE},i} \cdot f_i}
\label{eq:latency_final}
\end{equation}
This analytical model incorporates hardware-specific configurations, including process node scaling ($\alpha$) and processing efficiency ($\eta$), to enable device-level adjustment for accurate deployment simulation. The definitions of all variables are summarized in Table~\ref{tab:rnn_energy_latency_vars}.

\begin{figure*}
    \centering
    \includegraphics[width=0.90\textwidth]{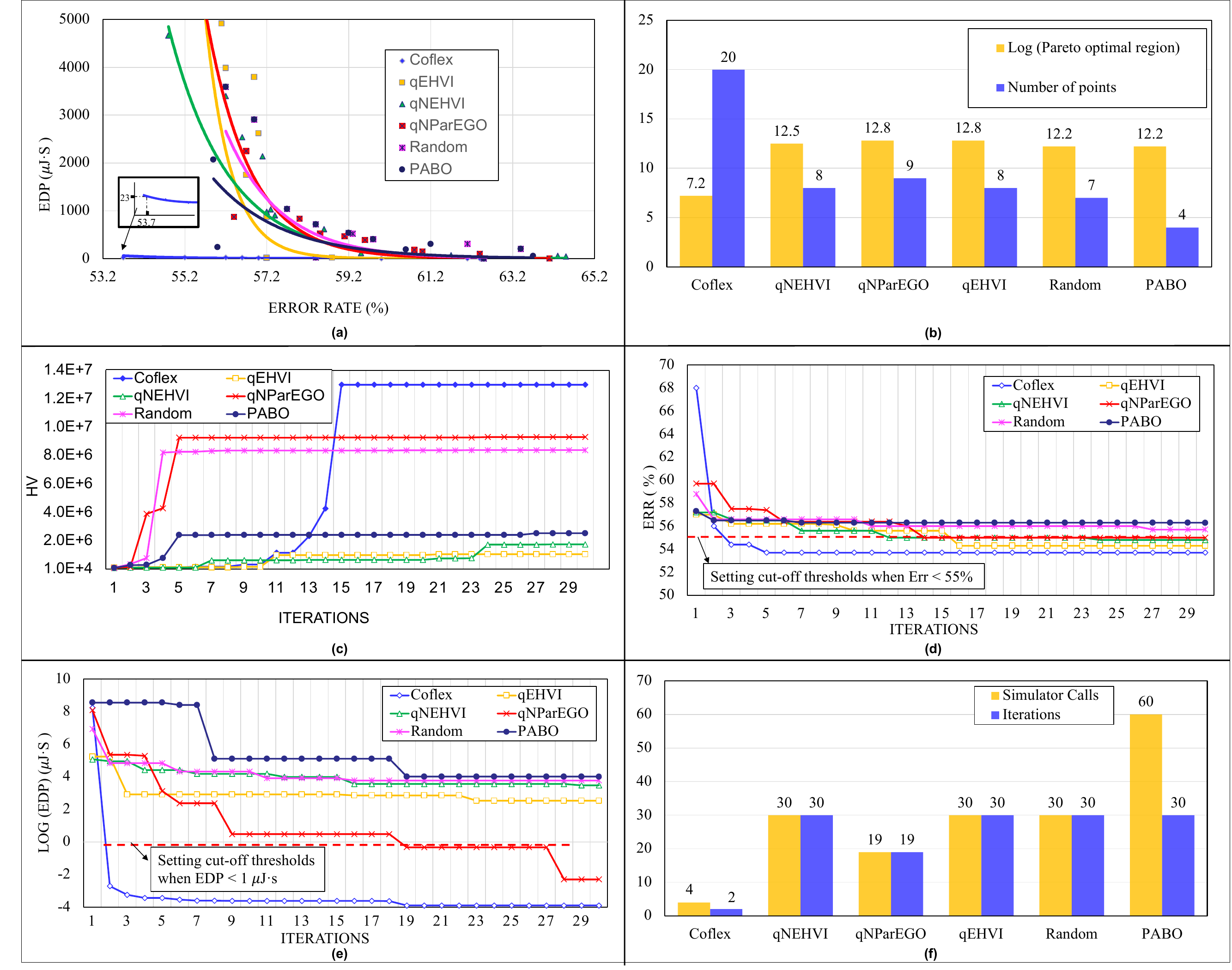}
    \caption{
    Performance Evaluation under Workload1 on ImageNet: (a) Pareto front in the EDP (uJ.s) vs error rate domain, illustrating the trade-off between hardware energy efficiency and prediction accuracy for Coflex, qNEHVI, qNParEGO, qEHVI, PABO, and Random. (b) Logarithm of the number of Pareto points, indicating convergence as the Pareto region shrinks. (c) Dominated Hypervolume progression, showing improved Pareto front quality over iterations. (d) Error rate convergence with star and diamond markers denoting when predefined thresholds are met. (e) Evolution of log(EDP), reflecting efficiency gains. (f) Resource analysis, presenting simulator calls and time cost (x10³) per iteration.
    }
    %{Performance Evaluation under Workload1 on ImageNet: (a) Pareto front in the EDP (uJ.s) vs error rate domain, illustrating the trade-off between hardware energy efficiency and prediction accuracy for Coflex, qNEHVI, qNParEGO, qEHVI, PABO, and Random. (b) Logarithm of the number of Pareto points, indicating convergence. (c) Dominated Hypervolume progression. (d) Error rate convergence. (e) Evolution of Log(EDP). (f) Resource analysis with simulator calls and time per iteration. \}
    \label{fig:comb_blks}
\end{figure*}

\section{Experimental Results}
\label{sec:experimental_results}

To compare with state-of-the-art HW-NAS methods, Coflex is evaluated on three distinct types of workloads (Table~\ref{tab:comparison-nas-bench}), each representing a different search space or application scenario.

\begin{enumerate}
    \item Type 1 (Image Classification):
    Utilizing NATS-Bench-SSS with a joint search space of $9.22\times10^{18}$ hyperparameters, we perform an assessment to examine whether Coflex can find solutions closer to the global optimum for image classification tasks over CIFAR10, CIFAR100, and ImageNet. 
    \item Type 2 (Semantic Segmentation): To evaluate Coflex on semantic segmentation, we conduct an experiment with TransNAS-Bench-101 and define a joint SW-HW search space of $1.15\times10^{18}$. The goal is to assess whether Coflex can maintain Pareto efficiency while handling high-resolution, multi-class output in semantic segmentation.
    % We test Coflex on TransNAS-Benchmark-101, which has $1.15\times10^{18}$ hyperparameters and covers various computer vision tasks, 
    %\textcolor{black}{Coflex shows strong generalization across semantic segmentation task, outperforming other SOTA methods on TransNAS-Bench-101.}
    %Coflex demonstrates superior Top-1 performance without framework modifications.
    \item Type 3 (\textcolor{black}{Natural} Language Processing): \textcolor{black}{We validate Coflex on sequence modeling tasks using NAS-Bench-NLP and define a joint SW-HW search space with $2.89\times10^{19}$ candidate configurations. The search space incorporates hyperparameters for the Transformer architecture and underlying hardware accelerator.}
    %This experiment targets language modeling under hardware constraints, examining Coflex’s ability to co-optimize both model perplexity and inference efficiency.} 
    % \item \textbf{Type 4 (Efficient Deployment on Edge Devices):} \textcolor{black}{To specifically evaluate performance under mobile deployment constraints, we apply Coflex on the DeFiNES simulator, which models energy and latency on resource-constrained Mobile-class CNN accelerators. While overall accuracy is lower compared to Type 1 due to stricter resource budgets, Coflex still achieves the Top-1 performance among all compared methods in this scenario, demonstrating its ability to generate Pareto-optimal SW-HW configurations that satisfy power-delay requirements without compromising relative accuracy.}
\end{enumerate}
To ensure fairness and reproducibility, all state-of-the-art benchmarks are compared under consistent datasets, task definitions, and hyperparameter configurations. We open-source the Coflex framework with a repository available at \url{https://github.com/Edge-AI-Acceleration-Lab/Coflex}

\subsection{Experimental Setup}
\label{sec:experiment_settings}

Tables~\ref{tab:exp-nas} and~\ref{tab:exp-sim} summarize the configuration details for type 1 workload, including hardware configurations and DNN settings. We compare Coflex with SOTA HW-NAS methods. \textcolor{black}{Specifically, when benchmarking with MOBO-based baseline~\cite{palonen_mobo_2013}, we employ a range of acquisition functions, including Random, qNEHVI~\cite{daulton2021parallel}, qEHVI~\cite{daulton2020differentiable}, and qNParEGO~\cite{knowles2006parego}. In addition, genetic-based algorithms such as NSGA-II~\cite{blank_pymoo_2020,deb2002fast} and agent-based strategies like PABO~\cite{parsa2020bayesian,parsa_pabo_2019} are included for comparison.}\footnote{%
Since PABO is not publicly available, we implemented our version based on the published pseudo-code, rigorously validating its accuracy through experimental verification to ensure minimal deviation from the original results.}
%In the same Case Study 1 setup, after 14 iterations, the obtained results include an error rate of 26.00\% and 7.89 mJ energy consumption, compared to the best reported result of 26.20\% and 8.12 mJ, indicating that our implementation achieves both lower error and energy consumption}

In this framework, \textcolor{black}{we utilize inference error rate as the software metric to show candidate neural network performance.}
%It exhibits a trade-off with hardware metrics in many workloads.}
%\textcolor{red}{validation error in training or testing error in inference?} 
%\textcolor{red}{We can not say so. Error rate is inherently important, regardless of hardware evaluation.}
\textcolor{black}{Besides, we adopt Energy Delay Product (EDP) as the hardware energy efficiency metric. Specifically, we obtain application-level energy consumption from the hardware simulator while getting latency numbers by taking the ratio of the total cycle count and the clock frequency that the accelerator is operating with.} 
%The EDP is therefore calculated as
%\begin{equation}
%\text{EDP} = \text{Energy} \times \text{Latency}
%\end{equation}
%where energy in $\mu$J and latency in seconds.

\subsection{Assessing Optimization Quality}
\label{assessing_optimization_quality_metrics}
\subsubsection{Top-1 Results}
\label{sec:top-1_results_comparison_across_multiple_algorithms}

The overall performance of Coflex is evaluated by calculating the Euclidean distance between the data pair of (\(error rate, EDP\)) and a fixed reference point (0,0). The software-hardware solution corresponding to the minimum distance is considered the top-1 result. 
% , which is defined as}
% \begin{equation}
%     \text{ref} = \left( \min_{i} \{\text{Error}_i\}, \min_{i} \{\text{EDP}_i\} \right)
% \label{eq:ref_point}
% \end{equation}
% \noindent
% where the minimums are taken over all observed candidate solutions. 
%\textcolor{red}{This is incorrect. If we take the min as a reference point, the optimal data pair will generate a distance of 0}

\subsubsection{Pareto-Front Filtering}
\label{sec:pareto-front_filtering}

\textcolor{black}{HW-NAS aims to explore the search space efficiently by jointly optimizing multiple objectives.} 
Besides Pareto front trade-offs, the performance of HW-NAS is also assessed through two metrics: the Pareto Optimal Region and the Dominated Hypervolume.
The Pareto Optimal Region (Figure~\ref{fig:comb_blks}(b)) is defined as the hypervolume between the reference point and the nearest Pareto front~\cite{10.1007/3-540-36970-8_38, rebello2021pareto}, which is a region indicating how close to the theoretical optimum the candidate solutions are. In contrast, the Dominated Hypervolume (Figure~\ref{fig:comb_blks}(c)) spans from the reference point to the furthest Pareto front~\cite{10.1007/BFb0056872, beume2007sms}. A larger volume indicates broader exploration and less risk of local optima. 
\textcolor{black}{
%These two metrics together provide a comprehensive assessment of optimization quality. 
An effective HW-NAS algorithm should both converge to a minimal Pareto Optimal Region and explore a broad solution space.}
%Such algorithms not only deliver competitive Top-1 accuracy but also improve overall solution quality across the Pareto front compared to SOTA methods.}

%\subsubsection{Optimization Efficiency Analysis}
%\label{sec:optimization_efficiency_analysis} 
\iffalse
HW-NAS algorithms for real-world deployment need to operate under hardware constraints such as latency requirements and power budgets. Therefore, optimization needs to rapidly explore the vast search space and identify models with high accuracy and low EDP, enabling local deployment and reducing cloud reliance.
\fi

\subsection{Result Analysis of Type 1 Workload}
\label{sec:wk_1}
In this experiment, input configurations from initial constrained sampling are mapped to the preset backbone network in NATS-Bench-SSS for image classification tasks.
\begin{table*}[ht]
\centering
\footnotesize
\caption{Comparison of HW-NAS methods on CIFAR-10, CIFAR-100, and ImageNet using NATS-Bench-SSS}
\label{tab:comparison-image}
\resizebox{1.0\textwidth}{!}{%
\begin{tabular}{l l l | ccc | ccc | ccc }
\toprule
\multicolumn{3}{c}{} & 
\multicolumn{3}{c}{\textbf{CIFAR-10}} & 
\multicolumn{3}{c}{\textbf{CIFAR-100}} & 
\multicolumn{3}{c}{\textbf{ImageNet}} \\
\cmidrule(lr){4-6}\cmidrule(lr){7-9}\cmidrule(lr){10-12}
\textbf{Search Space} & 
\textbf{Task} & 
\textbf{Method} & 
\textbf{Err(\%)} & 
\textbf{\texorpdfstring{EDP($\mu$J$\cdot$s)}{EDP (uJ·s)}} & 
\textbf{Dist.} & 
\textbf{Err(\%)} & 
\textbf{\texorpdfstring{EDP($\mu$J$\cdot$s)}{EDP (uJ·s)}} & 
\textbf{Dist.} & 
\textbf{Err(\%)} & 
\textbf{\texorpdfstring{EDP($\mu$J$\cdot$s)}{EDP (uJ·s)}} & 
\textbf{Dist.} \\
\midrule
\multirow{7}{*}{NATS-Benchmark-SSS} 
  & \multirow{7}{*}{Image Classification} 
  & \textbf{Coflex (SGP)} 
    & \textbf{9.29} & \textbf{0.95} & $\bm{5.52 \times 10^{-5}}$
    & \textbf{29.10} & \textbf{2.76} & $\bm{6.13 \times 10^{-4}}$
    & \textbf{53.70} & \textbf{23.00} & $\bm{2.87 \times 10^{-5}}$ \\[2mm]
  &  & MOBO(AF=qNParEGO) (GP)
    & 9.70 & 84.50 & 0.01 
    & 30.40 & 1.73 & 0.03 
    & 55.00 & 6710.00 & 0.04 \\[2mm]
  &  & MOBO(AF=qNEHVI) (GP)
    & 10.10 & 21.80 & 0.02 
    & 30.20 & 3.01 & 0.03 
    & 54.80 & 4670.00 & 0.04 \\[2mm]
  &  & MOBO(AF=qEHVI) (GP)
    & 10.10 & 53.60 & 0.02 
    & 30.50 & 16.50 & 0.04 
    & 54.30 & 6130.00 & 0.02 \\[2mm]
  &  & MOBO(AF=Random)
    & 10.20 & 35.70 & 0.02 
    & 31.00 & 25.50 & 0.05 
    & 56.00 & 15200.00 & 0.08 \\[2mm]
  &  & NSGA-II
    & 10.30 & 13.40 & 0.02 
    & 29.80 & 1.86 & 0.02 
    & N/A & N/A & N/A \\[2mm]
  &  & PABO (GP)
    & 10.20 & 17.60 & 0.02 
    & 30.50 & 97.10 & 0.04 
    & 55.90 & 2070.00 & 0.07 \\
\bottomrule
\end{tabular}%
}
\end{table*}
\begin{table*}[ht]
\centering
\footnotesize
\caption{\textcolor{black}{Comparison of Methods for Semantic Segmentation on Taskonomy using TransNAS-Bench-101}}
\label{tab:comparison-trans}
\resizebox{0.60\textwidth}{!}{%
\begin{tabular}{l l l | ccc }
\toprule
\multicolumn{3}{c}{} & \multicolumn{3}{c}{\textbf{Taskonomy}} \\
\cmidrule(lr){4-6}
\textbf{Search Space} & \textbf{Task} & \textbf{Method} & \textbf{Err(\%)} & \textbf{\texorpdfstring{EDP($\mu$J$\cdot$s)}{EDP (uJ·s)}} & \textbf{Dist.} \\
\midrule
\multirow{5}{*}{TransNAS-Bench-101} 
  & \multirow{5}{*}{Semantic Segmentation} 
  & \textbf{Coflex (SGP) } & \textbf{71.10} & \textbf{107.00} & $\bm{1.58 \times 10^{-3}}$ \\[2mm]
  &  & MOBO(AF=qNParEGO) (GP) & 71.30 & 586.00 & 0.01 \\[2mm]
  &  & MOBO(AF=qNEHVI) (GP) & 71.70 & 732.00 & 0.03 \\[2mm]
  &  & MOBO(AF=qEHVI) (GP) & 71.70 & 1320.00 & 0.03 \\[2mm]
  &  & MOBO(AF=Random) & 71.60 & 231.00 & 0.02 \\
\bottomrule
\end{tabular}%
}
% \begin{tablenotes}
%\item \textit{Note:} (SGP) indicates methods using Sparse Gaussian Process surrogates, (GP) refers to classical full-rank Gaussian Process models.
% \end{tablenotes}
\end{table*}
\begin{table*}[ht]
\centering
\footnotesize
\caption{\textcolor{black}{Comparison of Methods for Natural Language Processing on Penn Treebank + WikiText-2 using NAS-Bench-NLP}}
\label{tab:comparison-nlp}
\resizebox{0.60\textwidth}{!}{%
\begin{tabular}{l l l | ccc }
\toprule
\multicolumn{3}{c}{} & \multicolumn{3}{c}{\textbf{Penn Treebank + WikiText-2}} \\
\cmidrule(lr){4-6}
\textbf{Search Space} & \textbf{Task} & \textbf{Method} & \bm{$\log(Perplexity)$} & \textbf{\texorpdfstring{EDP($\mu$J$\cdot$s)}{EDP (uJ·s)}} & \textbf{Dist.} \\
\midrule
\multirow{5}{*}{NAS-Bench-NLP} 
  & \multirow{5}{*}{Language Modeling} 
  & \textbf{Coflex (SGP) } & \textbf{4.37} & $\bm{2.33 \times 10^{-5}}$ & $\bm{8.08 \times 10^{-8}}$ \\[2mm]
  &  & MOBO(AF=qNParEGO) (GP) & 4.39 & 0.03 & $\bm{2.87 \times 10^{-4}}$ \\[2mm]
  &  & MOBO(AF=qNEHVI) (GP) & 4.39 & 0.03 & $\bm{2.76 \times 10^{-4}}$ \\[2mm]
  &  & MOBO(AF=qEHVI) (GP) & 4.39 & 0.08 & $\bm{4.96 \times 10^{-4}}$ \\[2mm]
  &  & MOBO(AF=Random) & 4.37 & $\bm{2.68 \times 10^{-3}}$ & $\bm{1.54 \times 10^{-5}}$ \\
\bottomrule
\end{tabular}%
}
% \begin{tablenotes}
%\item \textit{Note:} (SGP) indicates methods using Sparse Gaussian Process surrogates, (GP) refers to classical full-rank Gaussian Process models.
% \end{tablenotes}
\end{table*}
%\noindent
%Co-optimization of neural network performance and hardware energy efficiency are then used to determine the top-1 results of the HW-NAS algorithms.
Table~\ref{tab:comparison-image} presents the results across CIFAR-10, CIFAR-100, and ImageNet classification. 
%Coflex outperforms baseline HW-NAS algorithms in both prediction accuracy and energy efficiency across diverse datasets. 
\textcolor{black}{Particularly on ImageNet, Coflex achieves an error rate of 53.70\% (i.e., an accuracy of 46.30\%) with an EDP of 23.00~$\mu$J$\cdot$s, lower than baselines across the evaluated datasets. Similar observations are made on CIFAR-100 and CIFAR-10, where Coflex attains lower error rates and reduced EDP values.} %\textcolor{red}{No need to make a short summary. Summary should appear in conclusion, abstract and introduction.} 
%To assess optimization efficiency, we evaluate each algorithm on ImageNet, CIFAR-10, and CIFAR-100, with a focus on the more challenging ImageNet task. 
Figure~\ref{fig:comb_blks}(d) and (e) 
%\textcolor{red}{these figures can be after Table IV.} 
illustrate the convergence over iterations. We record the number of iterations each method requires to reach the cut-off thresholds for error rate and EDP, respectively, and annotate the earliest successful iteration and convergence failure. Coflex reaches both thresholds within just two iterations, rapidly converging compared to all baseline methods. Particularly, qNEHVI, qEHVI, and Random search require more than 30 iterations to converge, while qNParEGO achieves convergence only after 19 iterations.

\textcolor{black}{Figure~\ref{fig:comb_blks}(f) exhibits that Coflex significantly reduces the number of simulator calls and optimization iterations. This substantially reduces the time cost as each simulator invocation incurs considerable computational overhead.} \textcolor{black}{Given the time cost, Coflex demonstrates a convergence speedup ranging from $1.5\times$ to $11.2\times$ in reaching predefined thresholds, compared to SOTA algorithms.}

\subsection{Result Analysis of Type 2 Workload}
\label{sec:wk_2}

This experiment evaluates semantic segmentation using the Taskonomy dataset from TransNAS-Benchmark-101, where software configurations are mapped to predefined architecture indices. According to the results shown in Table~\ref{tab:comparison-trans}, Coflex achieves an error rate of 71.10\% and an EDP of \SI{107.00}{}~$\mu$J$\cdot$s on the Taskonomy dataset. Particularly, on the Taskonomy dataset, the error rate is reduced by 0.28\% to 0.84\%, and the EDP is lowered by 53.68\% to 91.89\% compared to baseline algorithms.

\subsection{Result Analysis of Type 3 Workload}
\label{sec:wk_3}

%In this workload, the software configurations provided by the optimization algorithms correspond to architecture indices from NAS-Bench-NLP, 
Experimental results from each optimization algorithm are tested with NAS-Bench-NLP,
which offers predefined neural network architectures for language modeling tasks. The neural network performance of each neural network architecture is evaluated using \(Log(perplexity)\) as the primary metric on the Penn Treebank (PTB) and WikiText-2 datasets.
As Table~\ref{tab:comparison-nlp} shows, Coflex achieves a \(Log(perplexity)\) of 4.37 and an EDP of $2.33 \times 10^{-5}$~$\mu$J$\cdot$s, which outperforms SOTA baselines by 0.46\% and two orders of magnitude, respectively. 

\subsection{Runtime Analysis}
%\textcolor{red}{Insert your new figure on runtime with a brief analysis}

\begin{figure}[!t]
  \centering
  \includegraphics[width=\linewidth]{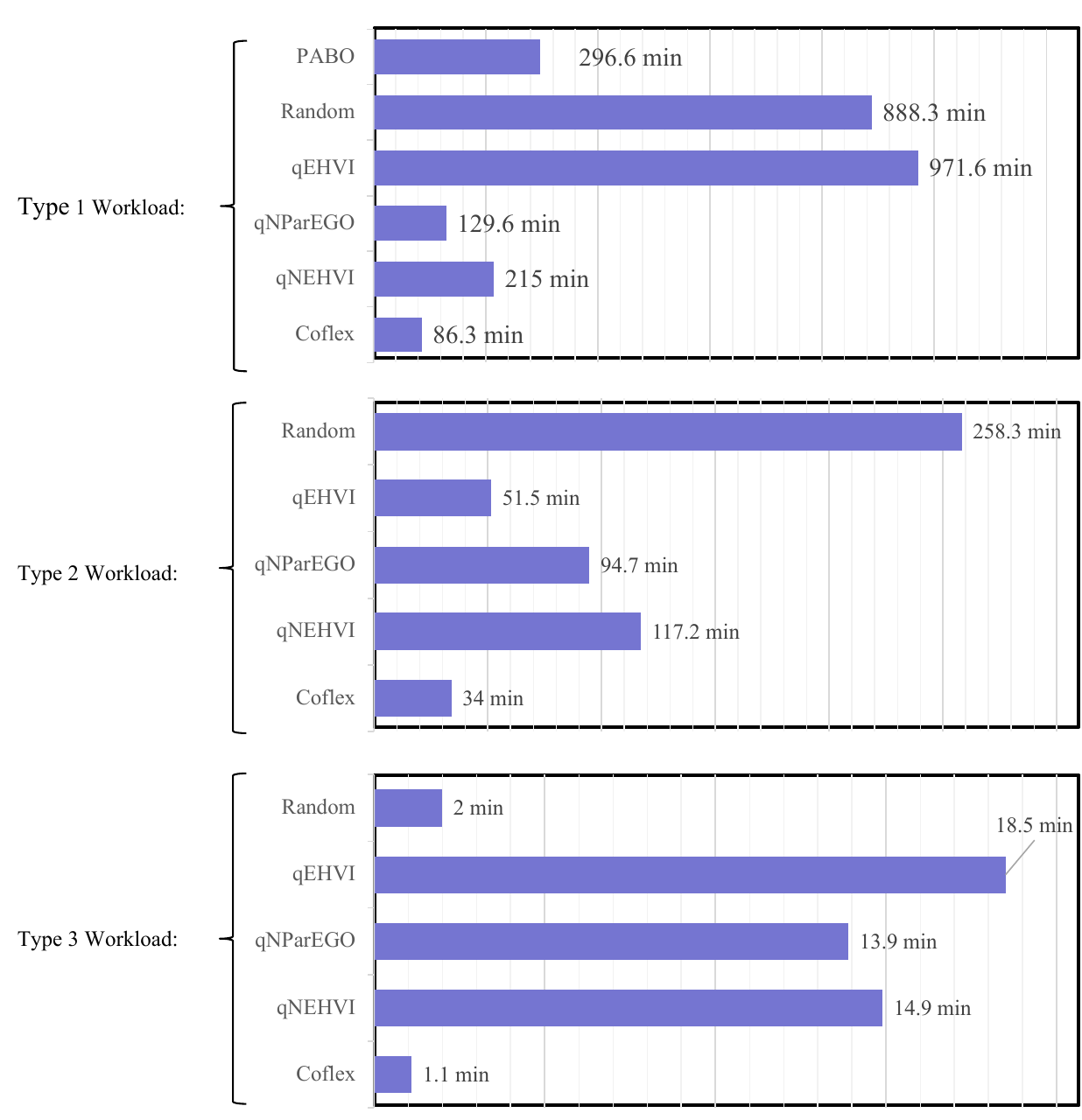}
  \caption{Comparison of runtime (in minutes) for different optimization methods across three workload types.}
  %Coflex demonstrates significantly reduced total optimization time across all cases.}
  \label{fig:runtime_comparison}
\end{figure}

Figure~\ref{fig:runtime_comparison} compares the runtime of Coflex with several baselines across three workloads. For type 1 workload, Coflex completes the task in 86.3 minutes, significantly outperforming qEHVI (971.6 min), Random Search (888.3 min), and PABO (296.6 min). Similar trends are observed in type 2 and type 3 workloads, where Coflex reduces the runtime to 34 minutes and 1.1 minutes, respectively. Notably, PABO fails to generate valid optimization results for the type-2 and type-3 tasks. This limitation is attributed to its pseudo-agent mechanism and reward design, which are tightly coupled with CNN-based classification pipelines. As a result, PABO lacks generalizability to benchmarks such as TransNAS-Bench-101 and NAS-Bench-NLP, which require support for dense prediction and sequence modeling architectures. In summary, our framework achieves 1.9× to 9.5× higher computational speed than the baselines across the three workloads.

%These results underscore Coflex's ability to optimize both architectural and hardware efficiency dimensions more effectively than baseline methods.

% \subsection{Workload 4: Efficient Deployment on Edge-TPU Devices \textcolor{red}{Suggest remove this part}}
% \label{sec:wk_4}

% \textcolor{black}{The objective in Workload~4 differs from Workload~1, focusing on evaluating lightweight NAS-Bench-201 architectures under Mobile-class edge-device deployment scenarios. \textcolor{red}{This belongs to type 1 workload, which is for image classification}
% This workload examines how well each HW-NAS algorithm adapts optimized neural architectures to Mobile-class processors, using the RBFlex-NAS framework to estimate software performance and the DeFiNES to estimate Mobile-class devices' hardware energy efficiency.}

% \textcolor{black}{As shown in Table~\ref{tab:comparison-image}, Coflex delivers outperformance results across all datasets. On ImageNet, Coflex achieves an error rate of 62.50\% (accuracy 37.50\%) and an EDP of 44.20~$\mu$J$\cdot$s, reducing error by 0.16\%–3.55\% and EDP by 8.10\%–92.63\% compared to baselines. Similar improvements are observed on CIFAR-10 and CIFAR-100. Coflex achieves 39.90\% error (60.10\% accuracy) and 54.40~$\mu$J$\cdot$s EDP, outperforming alternatives with error and EDP reductions of 4.51\% and 91.00\%, respectively.}

\section{Conclusion}

This work proposes Coflex, a Sparse Gaussian Process-based HW-NAS framework to facilitate large-scale software–hardware space exploration while maintaining low computational overhead. By leveraging sparse inducing points with Bayesian optimization, Coflex achieves significantly better top-1 results and Pareto-optimal trade-offs between accuracy and energy efficiency than existing state-of-the-art methods. 
%\textcolor{black}{On the NATS-Benchmark-SSS, Coflex achieved superior performance, attaining up to 4.10\% lower error rate and 99.85\% reduction in EDP compared to baseline methods.} 
\textcolor{black}{Moreover, Coflex exhibits faster convergence and substantially less runtime.}
\textcolor{black}{The results confirm that the Coflex framework enables efficient and scalable software-hardware co-design for edge DNN accelerators.}
% \bibliographystyle{IEEEtran}
% \bibliography{ref}

\end{document}